\documentclass[letterpaper, 10 pt, conference]{ieeeconf}  

\usepackage{amsmath,amssymb,amsfonts}
\usepackage[ruled,vlined]{algorithm2e}
\usepackage{algorithmic}
\usepackage{graphicx}
\usepackage{textcomp}
\usepackage{xcolor}
\usepackage{epsfig,color,ulem,url,booktabs}
\usepackage{multicol}
\usepackage{multirow}
\newsavebox{\tablebox}
\usepackage{subfigure}
\usepackage{comment}
\usepackage{ulem}
\newtheorem{defn}{{\bf Definition}}

\IEEEoverridecommandlockouts                              

\newcommand{\sig}[1]{{\small\textsf{{#1}}}}
\newcommand{\Comment}[1]{}
\newcommand{\oo}{\textbf{x}}

\title{\LARGE \bf
Monitoring Object Detection Abnormalities via Data-Label and Post-Algorithm Abstractions
}

\author{Yuhang Chen$^{1,2*}$, Chih-Hong Cheng$^{3*}$, Jun Yan$^{1,2,4}$ and Rongjie Yan$^{2,4}$%
\thanks{$^{*}$ The first two authors contributed equally to this work.}
\thanks{$^{1}$ Technology Center of Software Engineering, ISCAS, China}
\thanks{$^{2}$ University of Chinese Academy of Sciences,  China}
\thanks{$^{3}$ DENSO AUTOMOTIVE Deutschland GmbH, Germany} 
\thanks{$^{4}$ State Key Laboratory of Computer Science, ISCAS, China}
\thanks{Correspondence to yrj@ios.ac.cn, c.cheng@eu.denso.com}
\thanks{This project has received funding from the European Union’s Horizon 2020 research and innovation programme under grant agreement No 956123. }
}

\begin{document}

\maketitle
\thispagestyle{empty}
\pagestyle{empty}

\begin{abstract}

While object detection modules are essential functionalities for any autonomous vehicle, the performance of such modules that are implemented using deep neural networks can be, in many cases, unreliable. In this paper, we develop abstraction-based monitoring as a logical framework for filtering potentially erroneous detection results. 
Concretely, we consider two types of abstraction, namely data-label abstraction and post-algorithm abstraction. Operated on the training dataset, the construction of data-label abstraction iterates each input, aggregates region-wise information over its associated labels, and stores the vector under a finite history length. Post-algorithm abstraction builds an abstract transformer for the tracking algorithm. Elements being associated together by the abstract transformer can be checked against consistency over their original values. We have implemented the overall framework to a research prototype and validated it using publicly available object detection datasets.

\end{abstract}

\section{Introduction}

The safety of autonomous driving functions requires robust detection of surrounding objects such as pedestrians, cars, and bicycles. As implemented in the autonomous driving pipeline, the monitor function needs to report abnormal situations on the resulting detection. The filtering is challenging, as no labeled ground-truth in operation is offered to be compared. One common way of checking abnormality is to use other modalities to compare the result. For example, a monitor cross-checks the 3D bounding boxes generated from multiple cameras and 3D bounding boxes produced by other detection pipelines using LiDARs or radars.

The focus of this paper is to develop another type of monitor that does not exploit diversities in sensor modality. The consideration starts with a practical motivation where one also wants to achieve diversity in monitor design. Meanwhile, utilizing other sensor modalities can be further complicated by faults and component malfunctioning as introduced in the ISO~26262 context. Our key approach is to utilize the concept of \textit{abstraction}, where abstraction soundly aggregates all valuations that are considered legal. In contrast to non-abstraction methods (cf. Section~\ref{se:relatedwork} for related work) the benefit lies in the soundness inherently in the monitor. Moving beyond existing results in abstraction of neuron activation patterns~\cite{Cheng19,HenzingerL020}, we propose two new types of abstraction that are not based on extracting valuations of neurons for each image.

\begin{itemize}
    \item \textbf{(Data-Label Abstraction)}
    Data-label abstraction builds a summary over labels used in training, where abnormalities refer to the encountered input not contained in the abstraction. We present a general framework that builds the abstraction from $\lambda$-consecutive frames; for each frame, generate information based on the divided physical regions.

    \item \textbf{(Post-Algorithm Abstraction)} 
   Post-algorithm abstraction refers to methods that relax the immediate algorithm in the post-processing pipeline to filter infeasible values. We apply post-algorithm abstraction on the tracking algorithm that takes the input as bounding boxes from the object detection module.  Recall that standard tracking algorithms perform association based on objects with the same class due to the assumption of fault-absence in object detection; we demonstrate a simple post-algorithm abstraction technique that blurs the object class in the association step, while the association function is changed from 1-1 (in the standard algorithm) to 1-to-many (in the abstract transformer). Elements being associated together by the abstract transformer are then checked against consistency over their concrete, unabstracted values.

\end{itemize}

We have implemented the concept as a research prototype and evaluated the result using public autonomous driving datasets.
The implemented monitors find undesired cases such as objects in unexpected locations as well as objects with wrong classes. Applying a fine-tuned 
Faster R-CNN~\cite{ren2016faster} detector over  
the KITTI dataset~\cite{geiger2012we}, both types of monitors serve their purpose in filtering the problematic prediction without the need of human-in-the-loop with a success rate of~$84\%$. Simultaneously, the speed of abnormality filtering can reach the speed around~$20$ FPS, matching our minimum real-time requirement for deployment. 

The rest of the paper is structured as follows. After Section~\ref{se:relatedwork}
comparing related works with ours, Section~\ref{se:example} provides a motivating example to hint the underlying principle.
Section~\ref{se:concepts} presents required mathematical formulations such as regions. Sections~\ref{se:construction} and~\ref{se:postalg} provide details on  data-label abstraction and post-algorithm abstraction. Finally, Section~\ref{se:experimentation} demonstrates experimental results, and Section~\ref{se:conclusion} concludes the paper with future directions.

\section{Related work}\label{se:relatedwork}

In this work, we focus on monitoring methods without exploiting sensor modalities and diversities, where there exist also monitors equipped on the system level for fault checking and recovery~\cite{NFM20,xiang2021runtime}. Our attention is further restricted to monitoring modules implemented using learning-based approaches. An intuitive method to check abnormalities is to set a threshold on the output of the softmax function~\cite{hendrycks2017a}. 
Other techniques such as the temperature scaling technique can be applied to calibrate the confidence on final outputs~\cite{guo2017calibration}. The Monte-Carlo dropout technique~\cite{gal2016dropout,MC19} is an online method that utilizes dropout to create ensembles and to compute the Bayesian measure. The above methods are essentially confidence-based and require proper calibration in order to be used in safety-critical systems\footnote{The need for calibration is mentioned in clause 8.5.2. of the autonomous driving safety standard UL~4600~\cite{UL4600}.}. Currently, proper calibration with guaranteed error bounds is known to be hard for autonomous driving, as the data distribution for real-world driving is unknown and can be highly individualized. Even under the binary classification setup, the sharpness of calibrated confidence is hard to be guaranteed without prior knowledge of the distribution~\cite{gupta2020distribution}. Abstraction-based methods build a data manifold that encloses all values from the training dataset. Answering the problem of whether a newly encountered data point falls inside the manifold serves as a proxy of abnormality detection. Existing results in abstraction-based monitoring~\cite{Cheng19,HenzingerL020} build the manifold by extracting the feature vectors inside the neuron computation and are thus dependent on specific network configurations. Our work focus on two types of abstraction (data-label and post-algorithm) that are not directly related to the DNN being analyzed. For our post-algorithm abstraction techniques for monitoring, it is based on utilizing existing tracking algorithms~\cite{wojke2017simple,chen2018real}. However, tracking algorithms as specified in~\cite{wojke2017simple,chen2018real} assume that results from object detection are perfect in the predicted class, while our purpose is to filter imperfections in object detection. This brings a natural relaxation in our implementation, where our abstraction blurs the class association such that class flips can be efficiently filtered. However, the concept of post-algorithm abstraction is generic, and we expect it to be also applied to filter other types of errors for other modules in autonomous driving.

\section{A motivating  example}\label{se:example}
\begin{figure}[t]
	\centering
	\subfigure[Time $t-1$]{
		\begin{minipage}[t]{0.43\linewidth} 
\includegraphics[width=0.95\linewidth]{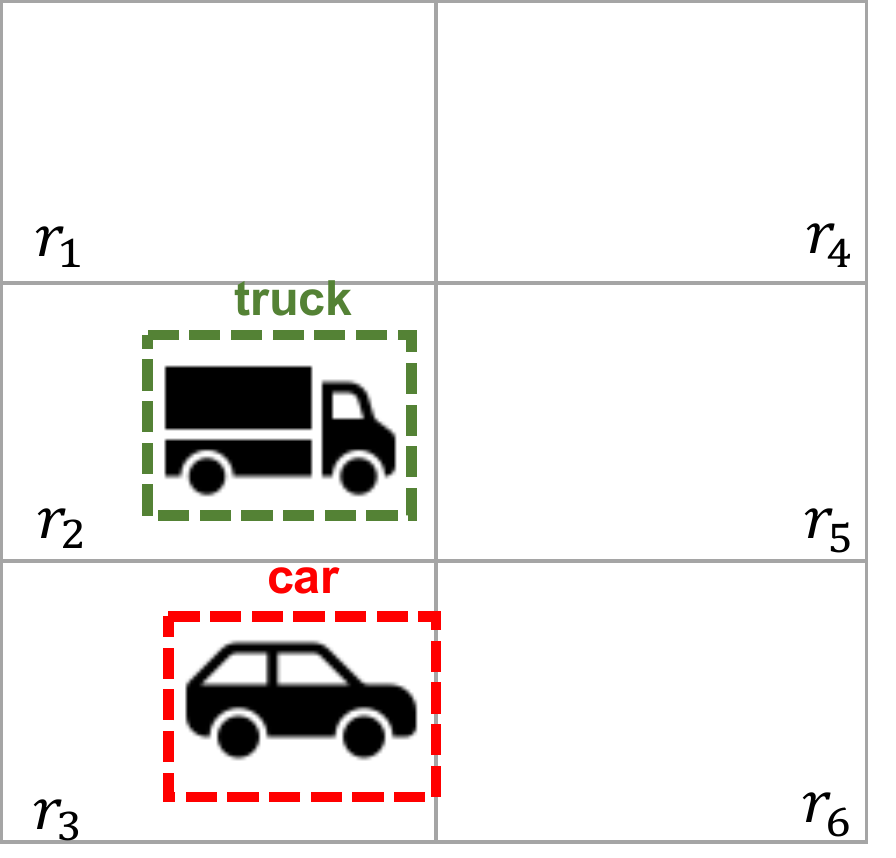}\label{fig:exmp1}
\end{minipage}
}%
\subfigure[Time $t$]{
		\begin{minipage}[t]{0.43\linewidth} 
\includegraphics[width=0.95\linewidth]{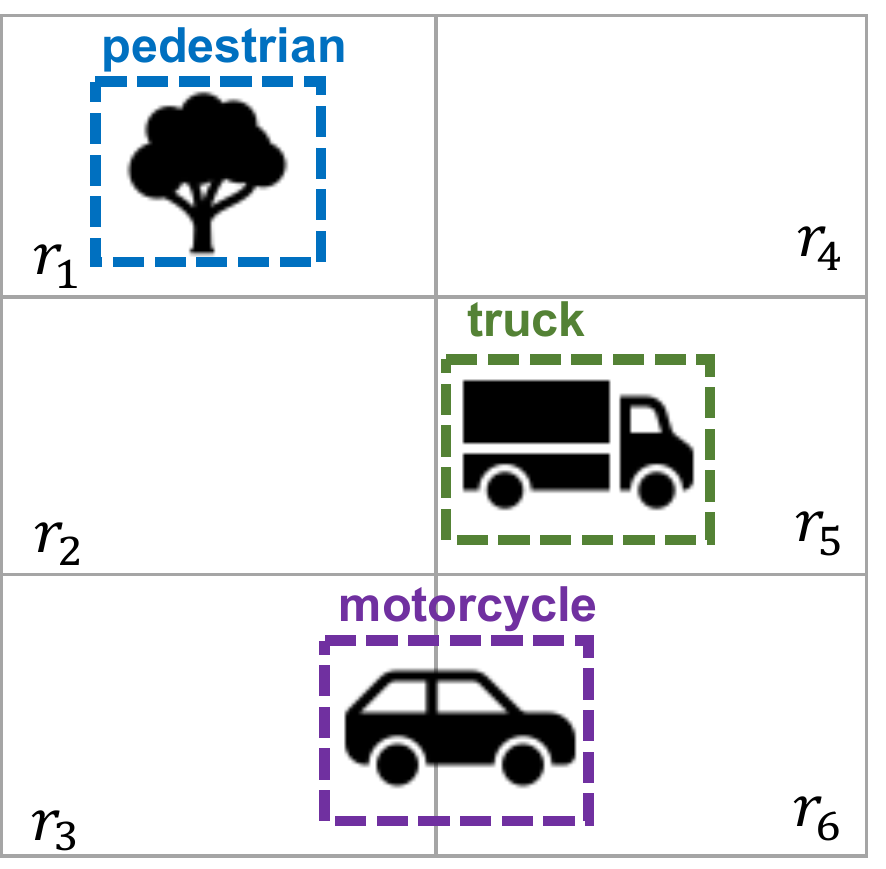}\label{fig:exmp2}
\end{minipage}
}
	\caption{A motivating example for abnormal checking}
	\label{fig:example}\vspace*{-\baselineskip}
\end{figure}

We present two consecutive images at timestamp~$t-1$ and~$t$ in Fig.~\ref{fig:example} to assist in understanding the concept of our monitors. The detected objects are marked with bounding boxes in dashed lines and the associated labels over the boxes. 
The monitor framework detailed in this paper will check two types of abnormalities at time~$t$:
\begin{itemize}
    \item The monitor based on \textit{data-label abstraction} finds that the identified pedestrian can be problematic, because in the training data, region $r_1$ never has a pedestrian.
    \item The monitor based on \textit{post-algorithm abstraction} finds that the motorcycle class can be problematic; by abstracting the information of object class, performing tracking, and then matching the tracked result, there seems to be a class flip over the detected object. 
\end{itemize}


\section{Preliminaries}\label{se:concepts}

We construct necessary notations to define the monitor mathematically. Here for simplicity, the formulation uses 2D object detection over the image plane, but the concept can be easily generalized to a 3D setup.

\begin{defn}[Region]
A region is an area denoted by $r=(pos_r, w,h)$, where $pos_r=(x,y)$ is the center point of the region, $w$ is the width of the region, and $h$ is the height of the region.
\end{defn}

An image is divided into a finite set of non-overlapping regions $R=\{r_i\}_{1\leq i \leq m}$. That is, given any $r_i$ and $r_j$ where $i\neq j$, their intersected area should be zero. 

We assume that the dataset contains images. For every image, its ground-truth label of an object $\oo=(class, pos,size)$ has three attributes: the label of the object, the localization $pos=(x,y)$ being the center point of the bounding box for the object in the image, and the size of the bounding box. Denote~$\mathbb{K}$ as the set of all possible classes.

Given two consecutive images, the physical locations of an object may shift. 
Then the movement of an object depicted from a sequence of images can be encoded as a finite trace.

\begin{defn}[Trace of an object]
For an object labeled in a sequence of $n$ consecutive images, let $(x_i,y_i)$ be the center point of the object at $i$-th image, and $size_i$ be its size. The trace of the object is denoted by $\tau=((x_1,y_1),size_1)\rightarrow \ldots\rightarrow((x_i,y_i),size_i)\ldots\rightarrow ((x_n,y_n),size_n)$, and we call  $\tau^i=((x_i,y_i),size_i)$ the $i$-th \textit{concrete state}. 
\end{defn}

Once an image is divided into regions, the location of an object can be mapped to a region based on the location of its center point. That is, an object $\oo$ is located in region~$r$, denoted by $\oo\odot r$, if the center point of the bounding box for  object $\oo$ is in the area covered by region~$r$. 
Therefore, a finite  trace of an object can be mapped to a region-based trace.

\begin{defn}[Region-based trace]
Given a trace of an object $\tau$ with $\tau^j=((x_j,y_j),size_j)$ and $1\leq j\leq n$, the \textit{region-based trace} is $\hat{\tau}=(r_{\langle 1 \rangle},size_1)\rightarrow \ldots \rightarrow (r_{\langle j \rangle},size_j) \rightarrow \ldots \rightarrow (r_{\langle n \rangle},size_n)$ such that $(x_j,y_j)$ is in the area covered by $r_{\langle j \rangle}$ and 
$r_{\langle j\rangle}\in R$.  The $j$-th \textit{(abstract) state} of a trace is denoted by $s^j=\hat{\tau}^j$, with $\hat{\tau}^j.region$ being the region of the state, and $\hat{\tau}^j.size$ being the size of the state.
\end{defn}

To simplify the notation, the region-based trace $\hat{\tau}$ of an object is written as a vector $((r_{\langle 1\rangle},size_1),\ldots$,$(r_{\langle n\rangle},size_n))$.

\begin{figure}[tp]
	\subfigure[Time $t-1$]{
		\begin{minipage}[t]{0.325\linewidth} 
\includegraphics[width=0.99\linewidth]{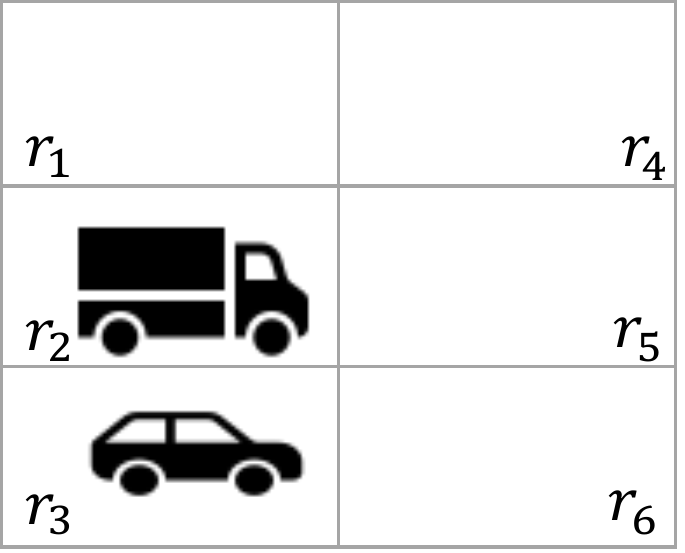}\label{fig:image1}
\end{minipage}\hspace{-5pt}
}%
\subfigure[Time $t$]{
		\begin{minipage}[t]{0.325\linewidth} 
\includegraphics[width=0.99\linewidth]{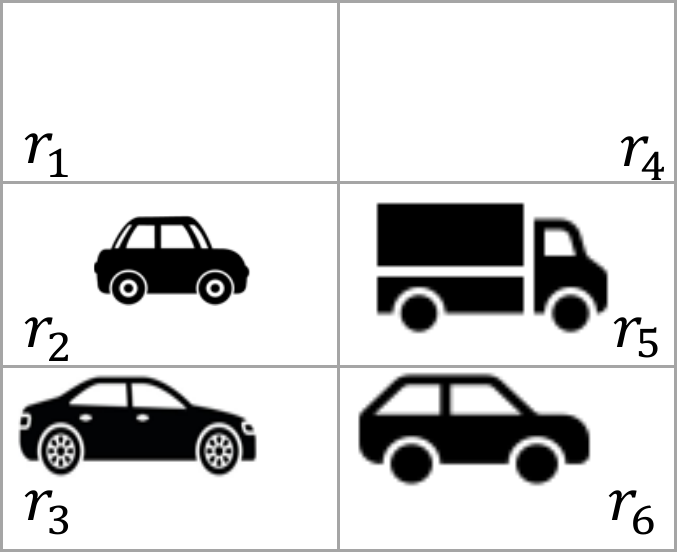}\label{fig:image2}
\end{minipage}\hspace{-5pt}
}%
\subfigure[Time $t+1$]{
		\begin{minipage}[t]{0.325\linewidth} 
\includegraphics[width=0.99\linewidth]{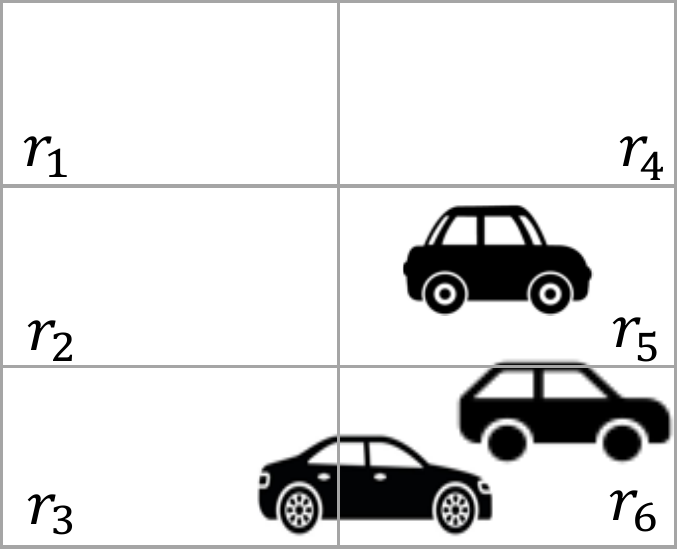}\label{fig:image3}
\end{minipage}
}%
	\caption{\vspace{-4pt}The ground-truth on a sequence of images}
	\label{fig:groundtruth}\vspace*{-\baselineskip}
\end{figure}

Consider the ground-truth on a sequence of consecutive images shown in Fig.~\ref{fig:groundtruth}.  Every image is divided into six regions. The set of classes on the objects involved in the sequence is $\{\sig{car}, \sig{truck}\}$. For the hatchback categorized as the 
\sig{car} object located in region~$r_3$ in Fig.~\ref{fig:image1}, it is also labeled in Figs.~\ref{fig:image2} and~\ref{fig:image3}. Therefore, its region-based trace is $((r_3,2.7),(r_6,3),(r_6,2.8))$ with the size of the vehicle changed from $2.7$ to $3$ to $2.8$ respectively. 
For the \sig{truck} located at $r_2$ in Fig.~\ref{fig:image1}, as it is not shown in Fig.~\ref{fig:image3}, its region-based trace is $((r_2,3),(r_5,3.5), (\bot, -1))$. We use $(\bot, -1)$ as a special region for \textit{undefined}, acting as a proxy of object disappearance in the image frame. 

\begin{defn}[$\lambda$-trace]
A $\lambda$-trace is a region-based trace, where the length of the trace is $\lambda$. Let $\mathcal{T}_{\lambda,k}$ be the set of $\lambda$-traces for objects in class $k$. 

\end{defn}

Reconsider the ground-truth shown in Fig.~\ref{fig:groundtruth}. The set of $2$-traces for the \sig{truck} class is $\mathcal{T}_{2, \sig{truck}}=\{((r_2,3),(r_5,3.5)), ((r_5,3.5),(\bot, -1))\}$. The set of $2$-traces for the \sig{car} class is 
\[\begin{array}{lll}
  \{((\bot,-1),(r_3,2.8)),& ((r_3,2.8),(r_6,2.9)), & //\textrm{big car}\\
((r_3,2.7),(r_6,3)), & ((r_6,3),(r_6,2.8)),&//\textrm{hatchback} \\  
((\bot,-1),(r_2,2)), &((r_2,2),(r_5,3))\}.  &//\textrm{small car}
\end{array}
\]

\section{Monitor construction on data-label abstraction}\label{se:construction}

The monitor on data-label abstraction can be regarded as a dictionary mapping from the set of classes $\mathbb{K}$ to an abstraction over the $\lambda$-traces accumulated from the available datasets.

\subsection{Monitor construction}
Let $D_k=\{((r_{\langle 1 \rangle}, [l_1,u_1]), \ldots, (r_{\langle \lambda \rangle}, [l_{\lambda},u_{\lambda}]))\}$ be the dictionary recording the information related to  $\lambda$-traces for class~$k$. The dictionary adopts interval $[l,u]$ to replace the attribute of \textit{size} in every state of a region-based trace, to record the size variation of the object in the region.

A $\lambda$-trace $\hat{\tau}\in \mathcal{T}_{\lambda,k}$ is \textit{strongly included} in $D_k$ if  $\exists \hat{\tau}_{\alpha}\in D_k$ such that  $\forall i\in [1, \lambda]:$ $(\hat{\tau}^i.region=\hat{\tau}^i_{\alpha}.region) \wedge (\hat{\tau}^i.size \in \hat{\tau}^i_{\alpha}.[l,u])$, denoted by $\sig{strcon}(\hat{\tau},D_k)$=\sig{true}. 
Additional to strongly inclusion, a $\lambda$-trace $\hat{\tau}\in \mathcal{T}_{\lambda,k}$ can be \textit{weakly included} in $D_k$, denoted by $\sig{wkcon}(\hat{\tau},D_k)$=\sig{true}, if $\sig{strcon}(\hat{\tau},D_k)$=\sig{false} and $\exists \hat{\tau}_{\alpha}\in D_k$ such that $\forall i\in [1, \lambda]: (\hat{\tau}^i.region=\hat{\tau}^i_{\alpha}.region)$, and $\exists j\in [1,\lambda]: \hat{\tau}^j.size \not\in \hat{\tau}^j_{\alpha}.[l,u])$. 

The construction of dictionary $D_k$ is depicted in Alg.~\ref{alg:labelbased}. 
The dictionary is updated w.r.t. the newly encountered or weakly included $\lambda$-traces. 
Given a newly encountered trace, before adding it to the dictionary, the size of every state in the trace is translated to an interval (e.g., from value~$5$ to interval $[5,5]$) by function \sig{size2inter()} (Lines~4,~5). 
Facing a weakly included trace (Line 11), the lower and upper bounds for the size intervals of the trace in the dictionary are expanded accordingly (Lines~14,~15). 
\begin{algorithm}[t]
\small
\caption{Construction on data-label abstraction}\label{alg:labelbased}
\begin{algorithmic}[1]
\STATE{Input: a set of $\lambda$-traces $\mathcal{T}_{k,\lambda}$ for objects of class $k$, $D_k$}
\STATE{Output: $D_k$}
    \FOR{($\forall \hat{\tau}\in \mathcal{T}_{k,\lambda}$)}
        \IF{ $D_k=\emptyset\vee \neg \sig{wkcon}(\hat{\tau}, D_k)$}
            \STATE{$D_k=D_k\cup \{\sig{size2inter}(\hat{\tau})\}$}
            \STATE{\textbf{continue}}
        \ENDIF
        \IF{($\sig{strcon}(\hat{\tau},D_k)$)}
                \STATE{\textbf{continue}}
        \ENDIF 
        \IF{($\sig{wkcon}(\hat{\tau},D_k))$}
                        
                        \STATE{Let $ \hat{\tau}_{\alpha}\in D_k$ where $\forall i \in [1, \lambda]: \hat{\tau}^i.region=\hat{\tau}^i_{\alpha}.region$}
                             \FOR{($j=1;j\leq \lambda; j++$)}
                                 \STATE{$\hat{\tau}^j_{\alpha}.l=\min(\hat{\tau}^j.size,\hat{\tau}^j_{\alpha}.l)$}
                                 \STATE{$\hat{\tau}^j_{\alpha}.u=\max(\hat{\tau}^j.size,\hat{\tau}^j_{\alpha}.u)$}

                            \ENDFOR

                \ENDIF
    \ENDFOR

\end{algorithmic}
\end{algorithm}

For the ground-truth presented in Fig.~\ref{fig:groundtruth}, the dictionary for the 2-traces  of class \sig{car} is constructed as follows:
\begin{enumerate}
    \item Initially, the dictionary is empty. The sizes of states $(\bot,-1)$ and $(r_3,2.8)$ in the first trace are replaced by the intervals $[-1,-1]$ and $[2.8,2.8]$, respectively. Then the trace is added. We have $D_{\sig{car}}=\{((\bot,[-1,-1]),(r_3,[2.8,2.8]))\}$. 
    \item In the second step, we cannot find the inclusion relation  between the trace in the dictionary and the one to be added. Therefore, after adding the second trace, the dictionary is 
    $\{((\bot,[-1$, $-1])$,$(r_3,[2.8,2.8])),((r_3,[2.8, 2.8]), (r_6,[2.9,2.9]))\}$.
    
    \item In the third step, the trace $((r_3,2.7),(r_6,3))$ is weakly included by the dictionary. Consequently, the intervals in the trace $((r_3,[2.8,2.8]),(r_6,[2.9,2.9]))$ are updated, and the dictionary is $\{((\bot,[-1,-1])$, $(r_3,[2.8,2.8])), ((r_3,[2.7,2.8]),(r_6,[2.9,3]))\}$.
    \item  When the process terminates, the  dictionary is \vspace{-6pt}
    \[\begin{array}{l}
    \resizebox{.9\hsize}{!}{$\{((\bot,[-1,-1]),(r_3,[2.8,2.8])),((r_3,[2.7,2.8]),  (r_6,[2.9,3]))$},\\
    \resizebox{.86\hsize}{!}{$((r_6,[3,3]),(r_6,[2.8,2.8])),((\bot,[-1, -1]),(r_2,[2,2]))$},\\
    \resizebox{.37\hsize}{!}{$((r_2,[2,2]), (r_5,[3,3]))\}$}.
    \end{array}
    \]

\end{enumerate}

\subsection{Abnormality checking}
To check abnormalities with the monitor on data-label abstraction, we need to compare the traces of the detected objects with the constructed dictionaries. 

Given a  $\lambda$-trace $\hat{\tau}$ of a detected object in class $k$, we check its inclusion relation with the dictionary $D_k$. 
To simplify the discussion, we assume that the first $\lambda-1$~states in the trace are normal. 
If the $\lambda$-trace $\hat{\tau}$ is not strongly included in $D_k$, i.e., $\sig{strcon}(\hat{\tau},D_k)=\sig{false}$, the detected result on the object is abnormal. Precisely, provided that $\sig{strcon}(\hat{\tau},D_k)= \sig{false}$, we may further refine the type of abnormality as follows: 
\begin{itemize}
    \item (Abnormal size) The abnormality occurs, if  $\sig{wkcon}(\hat{\tau},D_k)=\sig{true}$. 
    
    \item (Abnormal location) The abnormality occurs, if $\sig{wkcon}(\hat{\tau},D_k)=\sig{false}$.

    \item (Abnormal lost object) The abnormality occurs, if $\hat{\tau}^{\lambda}.region=\bot$ and 
    the condition of ``abnormal location" holds. In other words, if the detected $\lambda$-trace shows that the object is lost, and we cannot find such a trace in the dictionary, the loss on the object is abnormal. 

\end{itemize}

Consider the established dictionaries from the example in Fig.~\ref{fig:groundtruth}. Given a region-based trace $((r_3,2.8),(r_6,3.2))$ for the \sig{car} class, it is weakly contained by  dictionary $D_{\sig{car}}$, and the monitor reports \textit{abnormal size}. 
Given a region-based trace $((r_2,3),(r_4,2.8))$ for the  \sig{truck} class, region $r_4$ is not contained in the dictionary, and the location is abnormal.

\subsection{Robustness Considerations}\label{subsec.robustness}

\begin{figure}[tp]
\centering
\includegraphics[ width=0.85\linewidth]{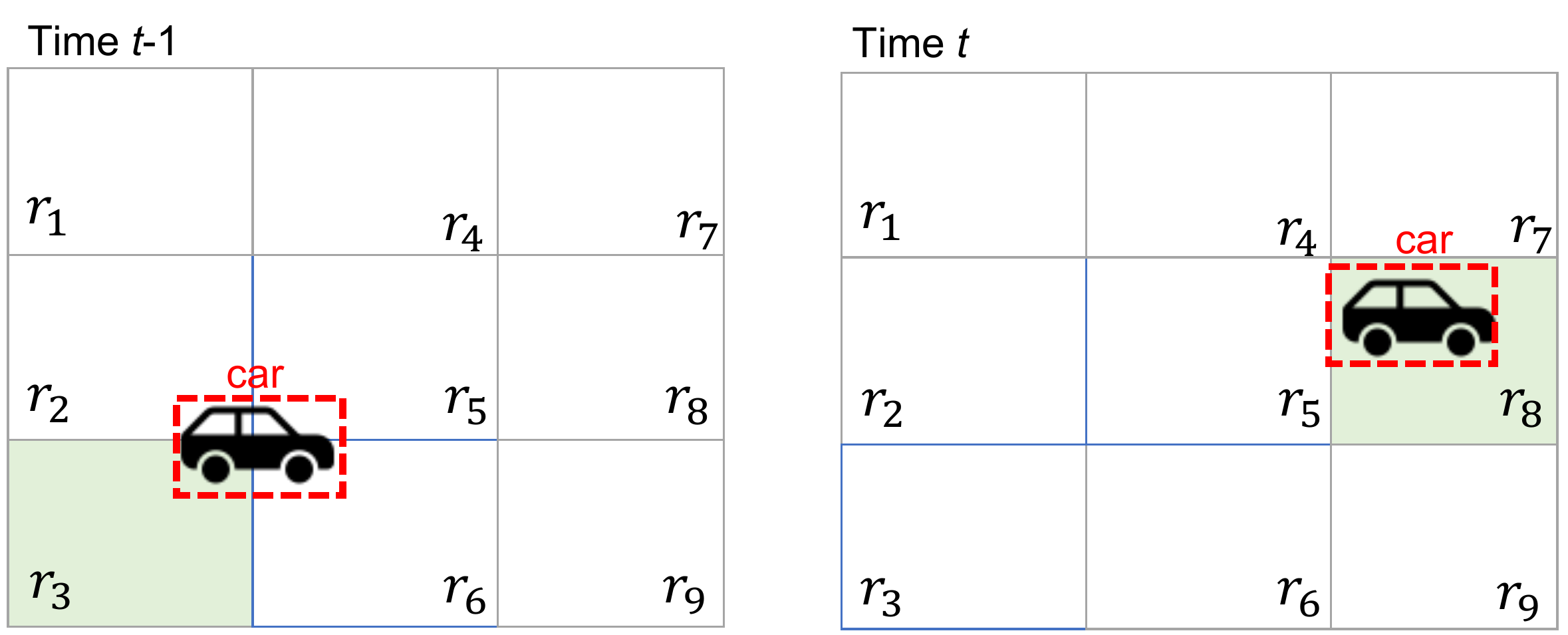}
	\caption{Robustness considerations in data-label abstraction}
	\label{fig:robustness}\vspace*{-\baselineskip}
\end{figure}

The method proposed in this section is based on  partitioning on the region of interest. In implementation, one commonly seen issue is related to the \textit{robustness} of the monitor caused by the region partitioning. An example can be seen in Fig.~\ref{fig:robustness}, where the center of the \sig{car} object at time $t-1$ is close to the top-right corner of region~$r_3$. Here we omit technical details, but under such cases, one can also consider $\delta$-ball perturbation on the center point of the bounding box. By doing so, additional three  $\lambda$-traces starting with region $r_2$, $r_5$ and $r_6$ will also be added in the monitor construction process, thereby providing \textit{robustness} guarantees on the monitor.


\section{Post-algorithm abstraction}\label{se:postalg}

\begin{figure}[htp]
\centering
	\subfigure[Detected results]{
		\begin{minipage}[t]{0.99\linewidth} 
\includegraphics[width=0.95\linewidth]{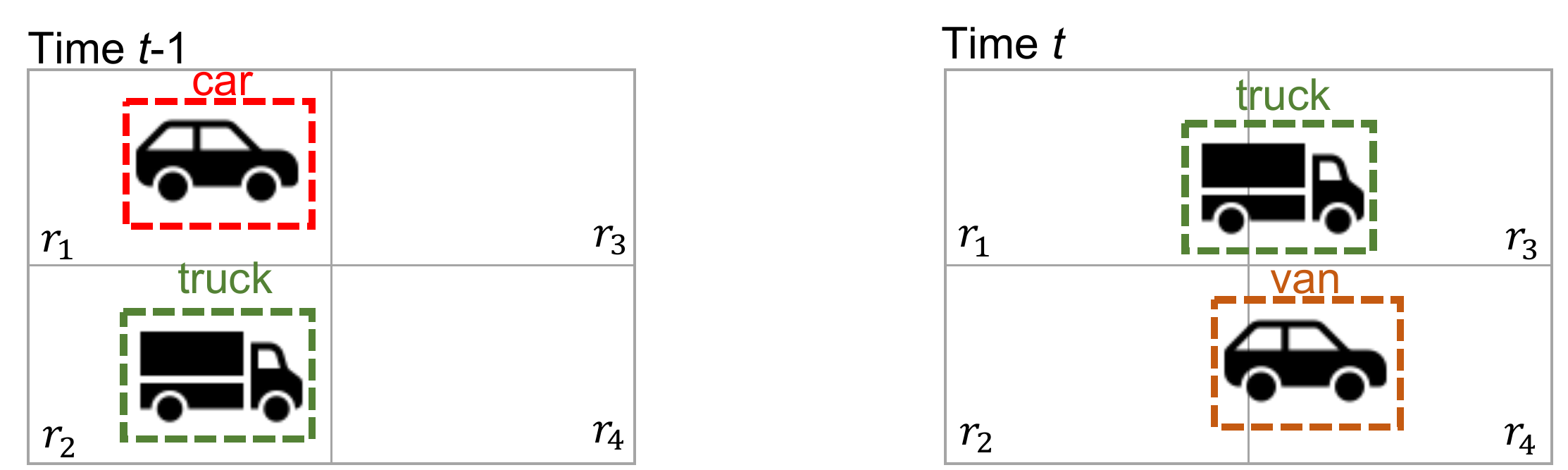}\label{fig:tracking1}
\end{minipage}
}\\
\subfigure[Post-algorithm abstraction -  class relaxation and correct association]{
		\begin{minipage}[t]{0.99\linewidth} 
\includegraphics[width=0.95\linewidth]{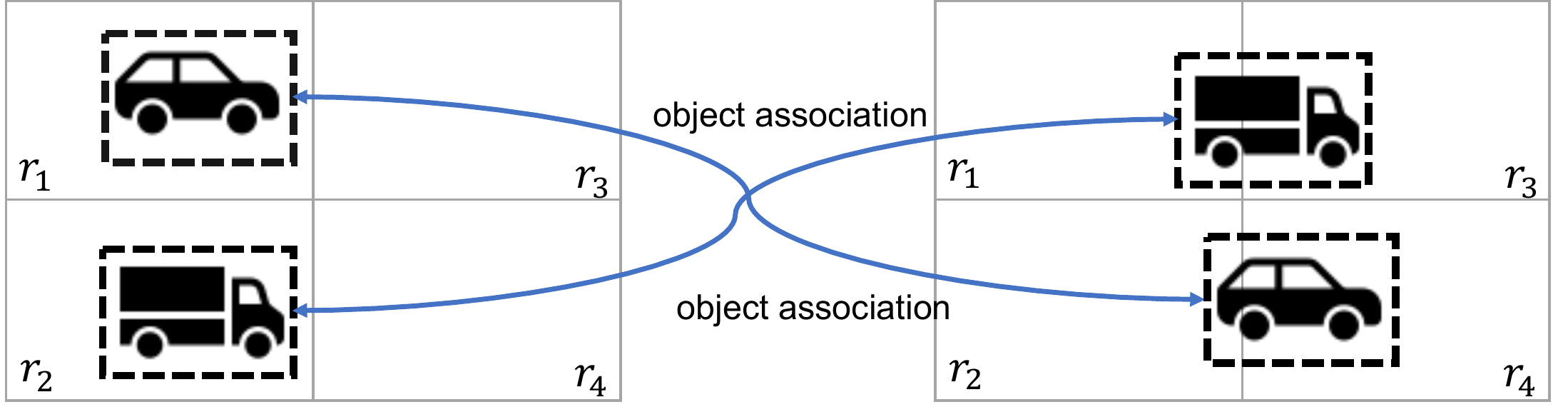}\label{fig:tracking2}
\end{minipage}
}\\
\subfigure[Post-algorithm abstraction -  class relaxation but incorrect association]{
		\begin{minipage}[t]{0.99\linewidth} 
\includegraphics[width=0.97\linewidth]{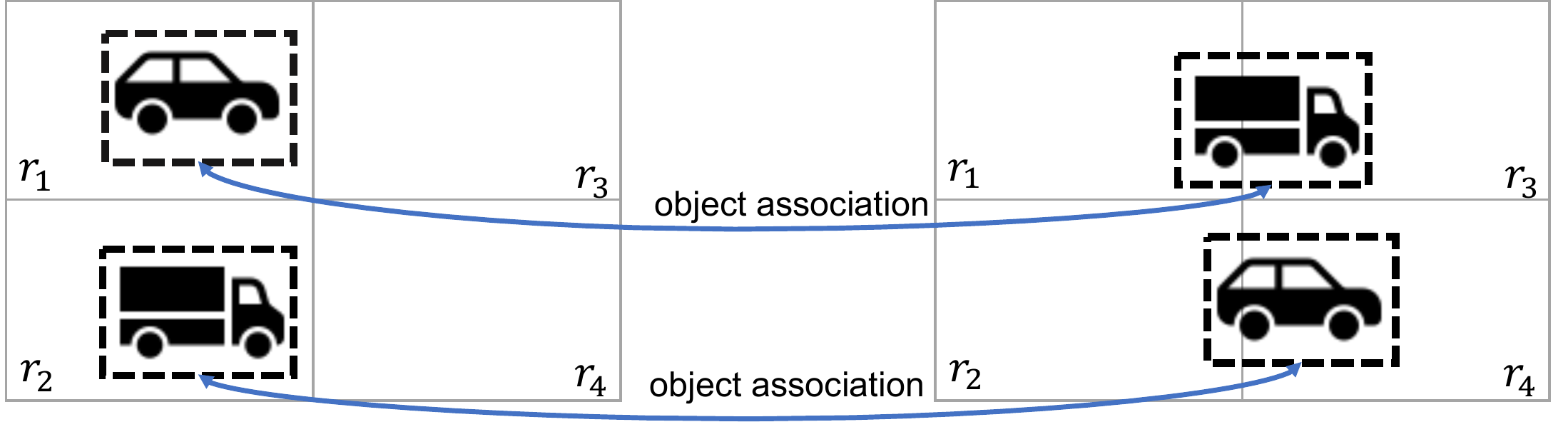}\label{fig:tracking3}
\end{minipage}
}\\
\subfigure[Post-algorithm abstraction -  class relaxation and weak association]{
		\begin{minipage}[t]{0.99\linewidth} 
\includegraphics[width=0.95\linewidth]{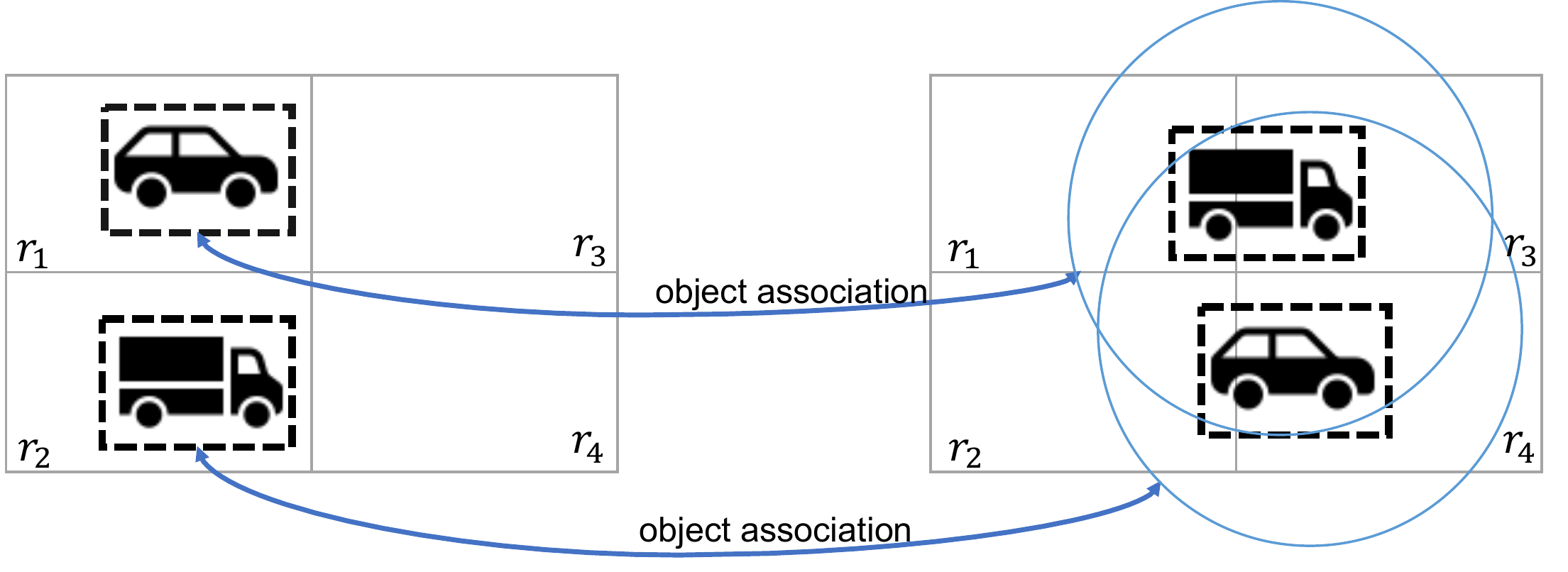}\label{fig:tracking4}
\end{minipage}
}%
	\caption{Understanding class relaxation and weak tracking}
	\label{fig:tracking}\vspace*{-\baselineskip}
\end{figure}

\subsection{Understanding the mechanism}

We start this section by providing an intuitive rationale on the process of post-algorithm abstraction using Fig.~\ref{fig:tracking}. Consider Fig.~\ref{fig:tracking1} being the result of object detection between two consecutive frames, where the vehicle identified as a \sig{car} object at time~$t-1$ is identified as a \sig{van} object at time~$t$. Therefore, we expect to have a monitor to filter the issue. To do so, we utilize the post-processing algorithm by generating an \textit{abstract transformer}. The abstract transformer performs an abstraction on the detected object such that the class information is blurred. Then perform object tracking which includes prediction and association. Provided that the tracking algorithm is correct similar to the case in Fig.~\ref{fig:tracking2}, it shall associate objects correctly. Then one can perform a \textit{concretization} step to restore the class label information, and compare against class labels. In Fig.~\ref{fig:tracking1}, the class flip of the \sig{car} object at time~$t$ will thus be detected. This is in contrast to a standard tracking algorithm that assumes correct class information, where standard tracking correctly relates the \sig{truck}, while reporting the disappearing of the \sig{car} and the appearing of the \sig{van}.

However, for the above-mentioned abstraction-based monitor to reduce false alarms, it is crucial to have correct object association between frames. This may be sometimes unrealistic due to objects being close by. Consider the association subroutine in the tracking algorithm producing incorrect results as shown in Fig.~\ref{fig:tracking3}. This leads to additional false alarms for the \sig{truck} object. Our mediation is to also \textit{relax the association} such that the association of an object at time~$t-1$ may be associated to multiple objects at time~$t$. This is to compensate the loss of precision due to blurring the class information. As demonstrated in Fig.~\ref{fig:tracking4}, the \sig{truck} object at time~$t-1$ will be weakly associated with the \sig{truck} object and the \sig{van} object at time~$t$, thereby no alarm will be reported. However, the \sig{car} object at time~$t-1$ will still be detected with  class flip. One can view the weak association in the post-algorithm abstraction monitor being analogous to the robustifying step for the data-label abstraction monitor highlighted in Section~\ref{subsec.robustness}.

\subsection{Monitor construction}
The construction of the monitor involves the following steps: 1) abstract or blur the classes of objects to be tracked; 2) perform the relaxed tracking over the objects without considering the classes of the objects; 3) enlarge the association candidates with a given bound; 4) restore the classes for the association candidate objects.

Let $\sig{hideclass(\oo)}$ be the function to abstract the class of object $\oo$, and the blurred object is $\hat{\oo}=\sig{hideclass(\oo)}$.
 Given an object $\oo^{t-1}$ at time~$t-1$, the tracked object with blurred class at time~$t$ is $\hat{\oo}^t=\sig{track}(\sig{hideclass}(\oo^{t-1}))$, where function \sig{track()} is the relaxed tracking algorithm without considering the class information of objects.

 We propose the concept of \textit{support set} as the set of potential tracked results for a given object.  
\begin{defn}[Support set]
Given an object $\oo^{t-1}$ at time~$t-1$, the set of candidates  specified by the post-algorithm abstraction is those being covered by a certain area, i.e., 
\begin{equation*}
     \resizebox{.99\hsize}{!}{$\sig{support}(\oo^{t-1})=\{\hat{\oo}^{t}\,|\, \hat{\oo}^{t}\odot \sig{radius}(\sig{track}(\sig{hideclass}(\oo^{t-1})),\delta)\}$},
\end{equation*}
where $\sig{radius}(\sig{track}(\sig{hideclass}(\oo^{t-1})), \delta)$ is the round area with the center point of the tracked object from  $\sig{track}(\sig{hideclass}(\oo^{t-1}))$ as the center and $\delta$ as the radius.  
\end{defn}

For example, for the image at time~$t$ shown in Fig.~\ref{fig:tracking4}, the support sets for the object in class \sig{car} and the object in class \sig{truck} at time~$t-1$ are the same, which are covered by the two  cycles in blue. 

\subsection{Abnormality checking}
The abnormality checking with post-algorithm abstraction is to check whether the class of an object at time~$t-1$ and the class of the tracked object at time~$t$ are consistent. 

To check the consistency, we need to restore the class information for the objects in the support set. Given an object $\oo^{t-1}$ at time~$t-1$, let $\sig{restoreclass}(\sig{support}(\oo^{t-1}))$ be the set of class labels for the potentially tracked objects from $\oo^{t-1}$  at time~$t-1$. If $\oo^{t-1}.class\not\in \sig{restoreclass}(\sig{support}(\oo^{t-1}))$, we say that a \textit{label flip} happens for the object $\oo$ at time~$t$.

As a final remark, the soundness of the post-algorithm abstraction relies on the tracking algorithm having a bounded error. In other words, if the tracking algorithm ensures that the predicted associated object only deviates from the real associated object with a distance of~$\delta$, the monitor can guarantee to detect class flips.

\section{Evaluation}\label{se:experimentation}

\begin{figure}[t]
	\centering

\subfigure[The traffic light on the left is mis-classified as \textsf{van}. The data-label abstraction monitor detects that it is impossible to have a \sig{van} of such a small size at that region.]{
		\begin{minipage}[t]{0.95\linewidth} 
\includegraphics[width=\linewidth]{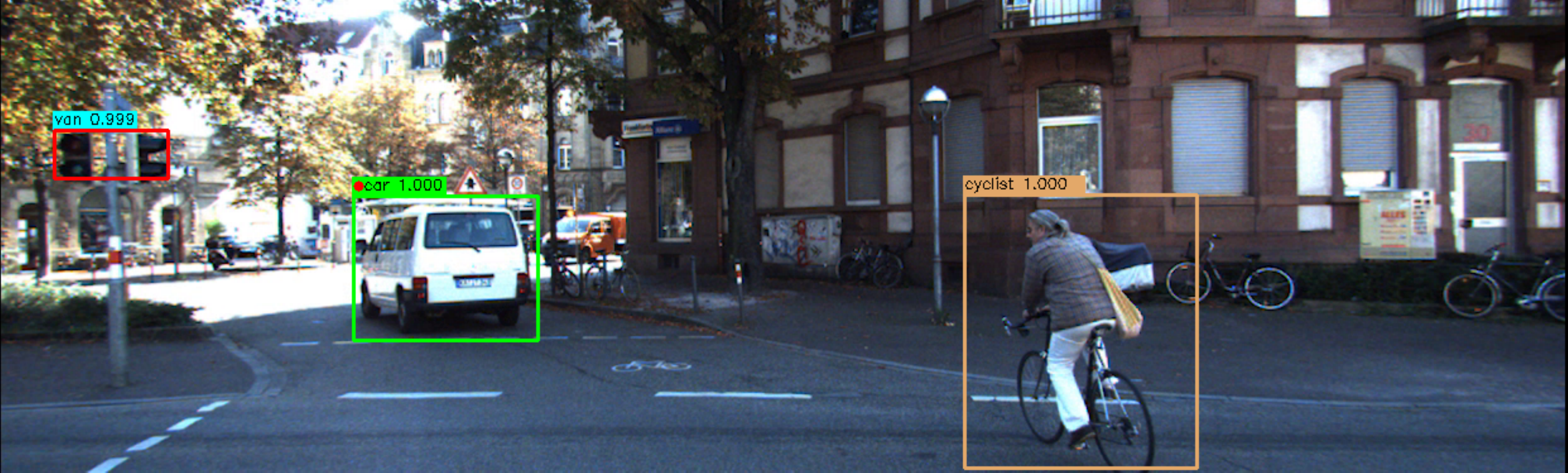}\label{fig:kitti1}
\end{minipage}
}\\
\vspace{-2mm}

	\subfigure[At time $t-1$, the bus object is detected as the \textsf{truck} class; in KITTI, buses and trucks share the same class. ]{
		\begin{minipage}[t]{0.95\linewidth} 
\includegraphics[width=\linewidth]{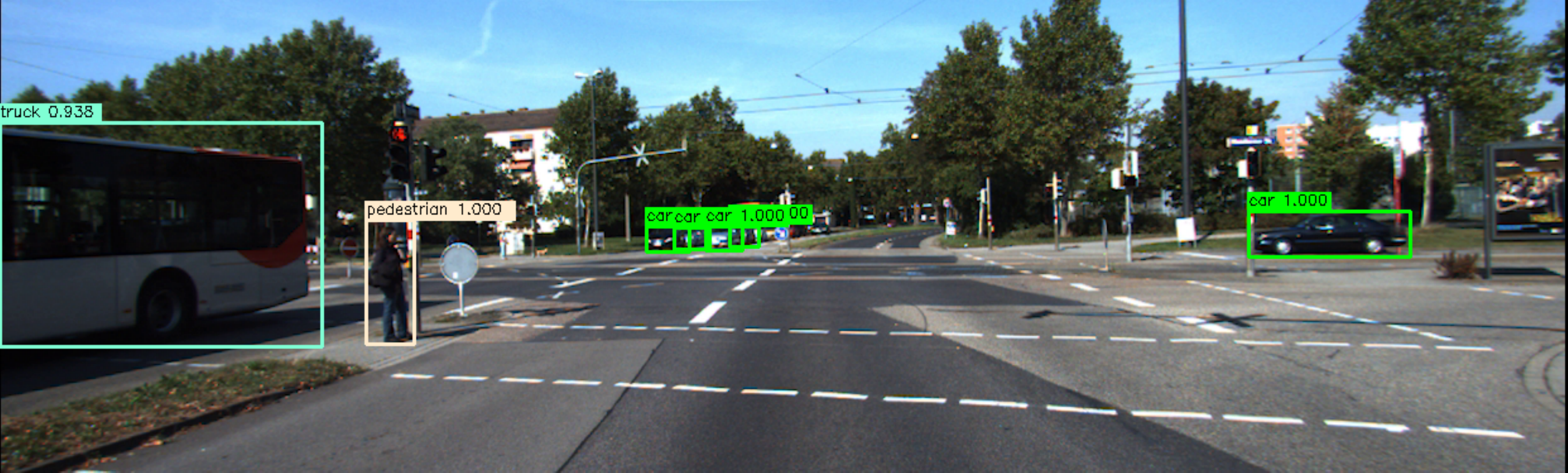}\label{fig:kitti2}
\end{minipage}
}\\%

\vspace{-2mm}

\subfigure[The post-algorithm abstraction monitor detects that the \textsf{truck} object at time $t-1$ is flipped to class \textsf{tram} at time $t$.]{
		\begin{minipage}[t]{0.95\linewidth} 
\includegraphics[width=\linewidth]{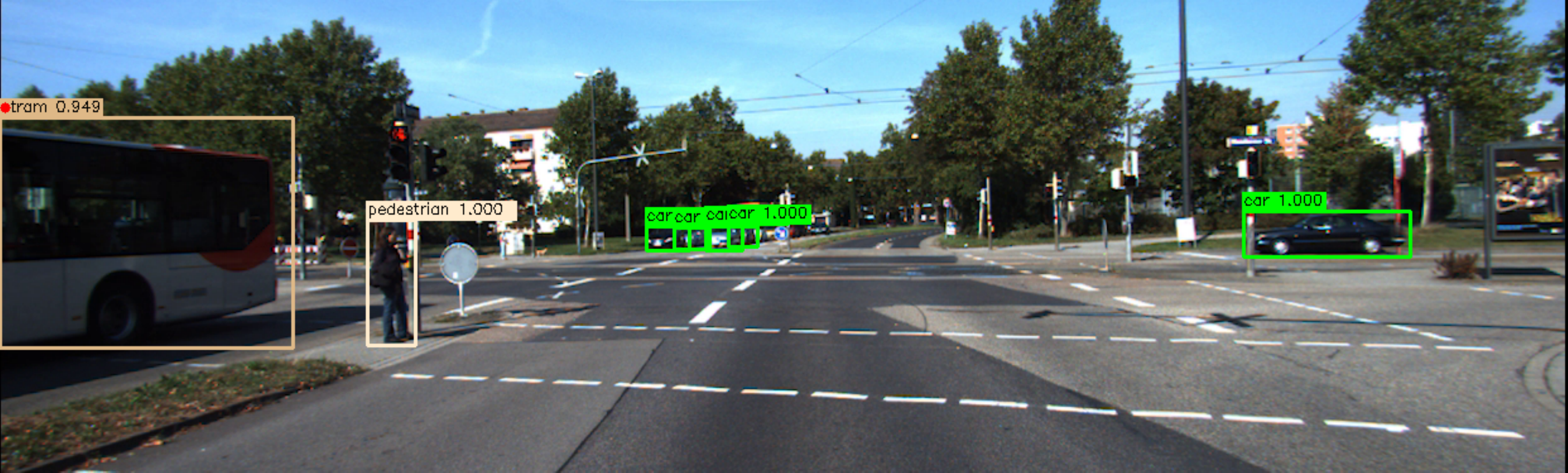}\label{fig:kitti3}
\end{minipage}
}

	\caption{Qualitative evaluation on the KITTI dataset}
	\label{fig:kitti}\vspace{-15pt}
\end{figure}

The prototype for the run-time monitor framework is implemented with Python, which currently supports  abnormality checking on \textit{abnormal location}, \textit{abnormal size}, \textit{abnormal object loss} and \textit{label flip}.
The experimentation runs on a PC equipped with Intel i7-7700@3.60GHz. We adopt eleven scenarios in the KITTI dataset\footnote{The evaluation using KITTI dataset in this paper is for knowledge dissemination and scientific publication and is not for commercial use.} for data-label abstraction. The other scenarios are adopted in validating the effectiveness of our method. For data-label abstraction, we set~$\lambda$ to~$1$ to record state-less behavior; for post-algorithm abstraction, we set~$\delta$ to~$0$ to avoid including other objects that are also nearby. Detecting ``abnormal object loss" equals to the case of setting $\lambda$ being~$2$ while the abstraction intuitively covers ``normal object loss" on two sides of an image (i.e., the object can disappear due to falling outside the image frame). The precision of the data-label abstraction monitor is highly dependent on the associated region size. For every image of size $1242 \times 375$ in the KITTI dataset, we build a total of $9\times 6 = 54$ regions. As explained in later subsections, the configuration offers a nice balance between computation speed and detectability. Fig.~\ref{fig:kitti} shows examples where errors are filtered by our implemented monitors.

\subsection{Effectiveness evaluation with ground-truth}
To evaluate the effectiveness of our run-time monitoring method, we randomly modify the labels and bounding boxes of labeled objects in the ground-truth of the dataset, and check whether the framework can find abnormalities. If an injected error is filtered, it is a \texttt{true positive} (TP). In other words, TP implies that there exists an error and monitor raises the warning. \texttt{false positive} (FP) implies that there exists no error and but monitor raises the warning.  We also adopt \texttt{recall} and  \texttt{precision} as the criteria for the evaluation. Table~\ref{tab:table1} provides a summary on the number of filtered abnormalities in each category, subject to the number of injected errors. From the table, we observe that the method can find all the injected label flips, with a slight cost of raising two additional false alarms. The false alarms are resulted from the setting of~$\delta$, which is not enough to cover close-by objects. For the other injected location errors, the monitor can find most of them. However, the precision for checking abnormal location is lower than those of the others. The reason is that the diversity of adopted scenarios for data-label abstraction is low, and some filtered abnormalities do not exist in those scenarios. The performance should be improved by applying more scenarios in various datasets in data-label abstraction.  

\begin{table}[t]
    \centering
    \caption{Effectiveness evaluation with ground-truth}\label{tab:table1}
     \begin{lrbox}{\tablebox}
    \begin{tabular}{|l| c| c|c|c|c|c|}\hline
        alarm type & $\#$ injected & TP & FP & precision & recall \\\hline
        abnormal location 
        & 34 & 27 & 5 & 0.844 & 0.794\\
        abnormal size & 50 & 47 & 6 & 0.887 & 0.940 \\
         object loss& 112 & 109 & 10 & 0.916 & 0.973 \\
        label flip & 180 & 180 & 2 & 0.989 & 1.000\\
        \hline
    \end{tabular}
     \end{lrbox}
\scalebox{0.95}{\usebox{\tablebox}}
\end{table}

\begin{table}[t]
    \centering
    \caption{Performance evaluation on run-time detected results}\label{tab:table2}
     \begin{lrbox}{\tablebox}
    \begin{tabular}{|c|l|c|c|c|c|c|}\hline
        scenario & alarm type & TP & FP & max OH & min OH & average\\\hline
        \multirow{4}{*} { 1} & abnormal location & 1 & 0 & & &\\
         & abnormal size & 7 & 1 & 0.155s & 0.025s & 0.064s\\
         &  object loss & 72 & 8 & & &\\
         & label flip & 78 & 4 & & & \\
         \hline
         \multirow{4}{*} { 2} & abnormal location & 1 & 0 & & &\\
         & abnormal size & 4 & 0 & 0.148s & 0.008s & 0.072s\\
         &  object loss& 157 & 29 & & &\\
         & label flip & 16 & 2 & & &\\
         \hline
         \multirow{4}{*} { 3} & abnormal location & 2 & 0 & & &\\
         & abnormal size & 4 & 1 & 0.126s & 0.001s & 0.058s\\
         &  object loss& 92 & 15 & & &\\
         & label flip & 15 & 2 & & &\\
         \hline
         \multirow{4}{*} { 4} & abnormal location & 0 & 0 & & &\\
         & abnormal size & 2 & 0 & 0.076s & 0.012s & 0.039s\\
         &  object loss& 25 & 2 & & &\\
         & label flip & 7 & 0 & & &\\
         \hline
         \multirow{4}{*} { 5} & wrong location & 0 & 0 & & &\\
         & unusual size & 25 & 6 & 0.130s & 0.012s & 0.047s\\
         &  object & 140 & 25 & & &\\
         & label flip & 37 & 6 & & &\\
         \hline
    \end{tabular}
    
     \end{lrbox}
\scalebox{0.9}{\usebox{\tablebox}}\vspace*{-\baselineskip}
\end{table}

\subsection{Evaluation on monitoring real-time detection results}
To evaluate its performance in monitoring real-time detection results, we collect the alarms and their categories and manually check their correctness. Meanwhile, we provide the run-time overhead (OH) to check whether the speed of the monitor is sufficient to meet the real-time requirement for self-driving. Table~\ref{tab:table2} summarizes the performance of our method in monitoring the detected results from 
fine-tuned Faster R-CNN~\cite{ren2016faster} detection network with Tensorflow Object Detection API~\cite{huang2017speed}. 
The selected scenarios involve various road conditions and traffic environments. 
Overall the average execution time is below $0.1$ second while some worst case can reach $0.155$ seconds. By careful examination on individual steps, it turns out that the weak-tracking is the overhead, as we relax a tracking algorithm utilizing histogram of gradients~\cite{dalal2005histograms}  and kernelized correlation filters~\cite{henriques2014high}.

Apart from further fine-tuning our implementations for faster monitoring speed, towards practical deployment with stricter real-time constraints, simple-to-achieve solutions may include either (1) adapting a simpler tracking algorithm or (2) simply reporting time-out when reaching the assigned budget and continue the next monitoring round.

\section{Concluding  Remarks}\label{se:conclusion}

In this paper, we considered the practical problem that the performance of object detection modules implemented with deep neural networks may not be reliable.  We developed abstraction-based monitoring as a logical framework for checking abnormalities over detected results. The data-label abstraction extracts characteristics of objects from existing datasets. The post-algorithm abstraction relaxes class labels in tracking algorithms to associate the objects in consecutive images. 
Our initial evaluation using publicly available object detection datasets demonstrated promise in integrating the developed technologies into autonomous driving products. 

The future work involves improving the performance of the prototype with more efficient tracking algorithms, strengthening the robustness of the framework with provable guarantees. Finally, as the concept of abstraction-based monitoring is a generic framework, we also plan to migrate the technique to filter other types of faults as well as move beyond object detection.

\bibliographystyle{IEEEtran}

\end{document}